\documentclass{article} 
\usepackage{iclr2022_conference,times}

\usepackage{amsmath,amsfonts,bm}

\def\eqref#1{equation~\ref{#1}}

\def\1{\bm{1}}

\DeclareMathAlphabet{\mathsfit}{\encodingdefault}{\sfdefault}{m}{sl}
\SetMathAlphabet{\mathsfit}{bold}{\encodingdefault}{\sfdefault}{bx}{n}
\newcommand{\tens}[1]{\bm{\mathsfit{#1}}}

\def\tP{{\tens{P}}}

\def\sL{{\mathbb{L}}}

\def\sX{{\mathbb{X}}}

\usepackage{hyperref}
\usepackage{url}
\usepackage{algorithm}
\usepackage{algorithmic}
\usepackage{booktabs}       
\usepackage[pdftex]{graphicx}       
\usepackage{subcaption}     
\usepackage{placeins}

\usepackage{wrapfig}

\title{Feature Flow Regularization: Improving Structured  Sparsity in Deep Neural Networks}

\author{%
  Yue Wu \\
  Department of Mathematics\\
  HKUST\\
  \texttt{ywudb@connect.ust.hk}\\
  \And
  Yuan Lan \\
  Department of Mathematics\\
  HKUST\\
  \texttt{ylanaa@connect.ust.hk} \\
  \And
  Luchan Zhang \\
  Department of Mathematics\\
  HKUST\\
  \texttt{malczhang@ust.hk}\\
  \And
  Yang Xiang \\
  Department of Mathematics\\
  HKUST\\
  \texttt{maxiang@ust.hk} \\
}

\iclrfinalcopy 
\begin{document}

\maketitle

\begin{abstract}
Pruning is a model compression method that removes redundant parameters and accelerates the inference speed of deep neural networks (DNNs) while maintaining accuracy. Most available pruning methods impose various conditions on parameters or features directly. In this paper, we propose a simple and effective regularization strategy to improve the structured sparsity and structured pruning in DNNs from a new perspective of evolution of features. In particular, we consider the trajectories connecting features of adjacent hidden layers, namely feature flow. We propose feature flow regularization (FFR) to penalize the length and the total absolute curvature of the trajectories, which implicitly increases the structured sparsity of the parameters. The principle behind FFR is that short and straight trajectories will lead to an efficient network that avoids redundant parameters. Experiments on CIFAR-10 and ImageNet datasets show that FFR improves structured sparsity and achieves pruning results comparable to or even better than those state-of-the-art methods.
\end{abstract}

\section{Introduction}
\label{sec: introduction}
Deep neural networks (DNNs) have achieved huge success in a wide range of applications.
Meanwhile, DNNs require considerably more computational cost and storage space as they become deeper in order to achieve higher accuracy.
\cite{Denil} demonstrated that there is significant redundancy in the parameterization of DNNs.
The Lottery Ticket Hypothesis \citep{LTH}
conjectures that there exist sparse sub-networks that can obtain a comparable accuracy with the original network when trained in isolation.

Model compression methods have been proposed to balance accuracy and model complexity,
e.g. weight pruning \citep{LeCun, Hassibi, Han, Guo, Trim, Li_filter} and quantization \citep{quantization},
low-rank approximation \citep{Denton, Jaderberg, Liu},
and sparsity structure learning \citep{Wen, Jose, Zhou, NSlimming, L0, VACL, SCS}.
Weight pruning removes less important parameters in the network.
In particular, filter pruning \citep{Li_filter} removes entire filters in the network together with their related channels, which can compress and accelerate DNNs efficiently.

Existing structured pruning methods can be divided into two categories: parameter-based methods
\citep{Li_filter, Pruning_retl, NSlimming, he2018soft, GAL, FPGM, Polarization, Provable_FP} that use some criteria to identify unimportant filters and remove them,
and feature-based methods \citep{ThiNet, SunJian, DCP, SMLI, FeatureS, hrank}
that select unimportant feature maps and then remove related filters and channels.
For example,
\cite{FeatureS} incorporated two feature map selections: discovering features with low diversity and removing features that have high similarities with others.

In this paper, we propose a new regularization method on the trajectory connecting features of adjacent hidden layers, namely feature flow regularization (FFR).
FFR smooths the trajectory of features, which implicitly improves the structured sparsity in DNN.
Our motivation is that the trajectory of data along the network reflects the DNN structure.
Shorter and straighter trajectory corresponds to an efficient and sparse structure of DNN.
An illustration is given in Figure \ref{fig: 01}b.

Our main contributions are:
(1)
We propose a new regularization (FFR) on the trajectory connecting the features of hidden layers, to improve the structured sparsity in DNN
from a perspective of the trajectory of data along the network.
This method is different from the existing sparsity structure learning methods, which directly impose regularization or constraints on the parameters.
Our method is also different from those pruning methods based on feature maps, which use the information of the feature map individually or in pairs (for similarity) without global relationship.
(2) We analyze the effect of FFR applied to convolutional layer and residual layer, and show that
FFR encourages DNN to learn a sparse structure during training by penalizing the sparsity of both parameters and features.
(3)
Experimental results show that FFR achieves a comparable or even better pruning ratio in terms of parameters and FLOPs than recent state-of-the-art pruning methods.

\FloatBarrier
\section{Related work}

\paragraph{Filter pruning.}
Various criteria for filter selection in pruning have been proposed.
\cite{Li_filter} used $L_1$ norm to select unimportant filters and
removed the filters whose norm is lower than the given threshold together with their connecting feature maps.
\cite{Pruning_retl} measured the importance of filters based on the change in the cost function induced by pruning.
\cite{ThiNet, SunJian} formulated pruning as a constraint optimization problem and selected most representative neurons based on minimizing the reconstitution error.
\cite{GAL} pruned filters as well as other structures by generative adversarial learning.
\cite{FPGM} pruned redundant filters utilizing geometric correlation among filters in the same layer.
\cite{Trim, DCP, FeatureS, hrank} removed filters based on the information, e.g. sparsity, rank or diversity, of feature maps that are generated by the filters.
Our method learns sparse DNN during training and adopts magnitude-based pruning scheme after training.

\paragraph{Sparsity regularization.}
Some studies introduced sparsity regularization to find sparse structure of DNN.
A commonly used strategy is to impose group Lasso and relaxed $L_0$ regularization \citep{Zhou, Wen, Jose, L0}.
\cite{NSlimming, Huang} associated a scaling factor with feature maps and imposed regularization
on these scaling factors during training to automatically identify unimportant channels. The feature maps with small scaling factor values will be pruned.
\cite{VACL} imposed a Cross-Layer grouping and a Variance Aware regularization on the parameters to improve the structured sparsity for residual models.
\cite{Polarization} used polarization regularizer on scaling factors.
 \cite{SCS} proposed a method to dynamically grow deep networks by continuously sparsifying structured parameter sets.
Different from these available methods that directly introduce regularization on the parameters or scaling factors,
our FFR imposes regularization on the trajectory connecting features of hidden layers to control the parameters and enforce structured sparsity implicitly.

\section{Method}

\subsection{Feature flow regularization}\label{sec: ffr}

We define \textbf{feature flow} as the trajectory formed by connecting the output features of adjacent hidden layers. For a DNN, the collection of trajectories with different input data from the training dataset reflect the network structure. We control these trajectories to obtain a sparse network.

\begin{figure}[!htb]
\centering
\begin{tabular}{cc}
    \begin{minipage}{0.37\textwidth}
    \centering
    \includegraphics[width=1.0\textwidth]{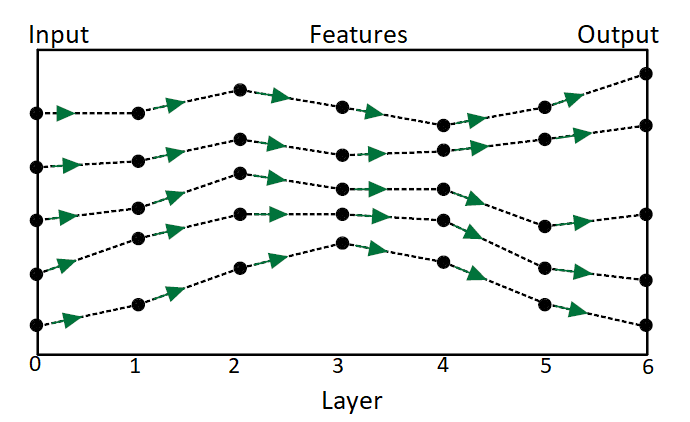}
	\subcaption*{(a)\label{fig: featureflow}}
    \includegraphics[width=0.9\textwidth]{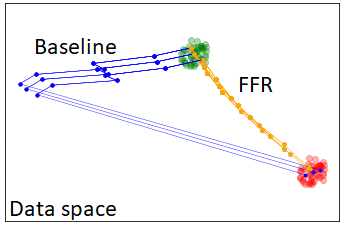}
    \subcaption*{(c)\label{fig: ffr_effect}}
    \end{minipage}&
    \begin{minipage}{0.58\textwidth}
    \centering
    \includegraphics[width=1.0\textwidth]{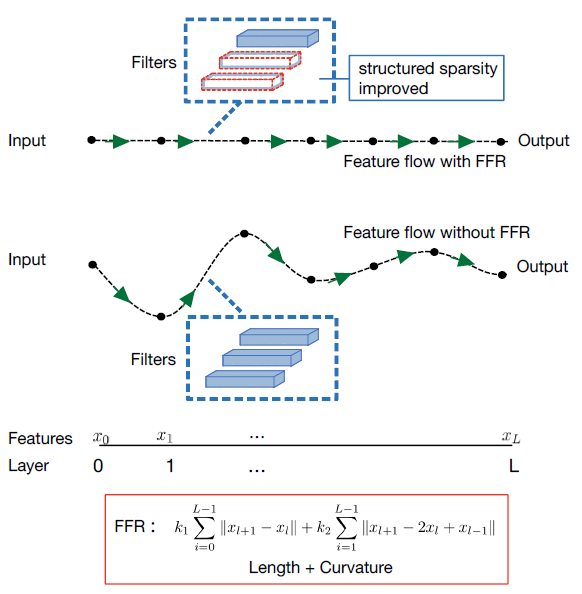}
	\subcaption*{(b)\label{fig: ff_prune}}
    \end{minipage}
\end{tabular}
\caption{(a) Feature flow demonstration: Each curve represents a feature flow, i.e., the trajectory connecting the features of hidden layers, where nodes are input, features and output.
    (b) Feature flow regularization demonstration: Feature flow under FFR is shorter and straighter due to the length and curvature penalty. As a result, FFR improves the structured sparsity and leads to effective filter pruning.  
    (c) An illustration example in two dimensional space showing the smooth effect of FFR, which is a five-block ResNet trained with FFR and without FFR (the baseline).
     The input data, features and targets are all points in two dimensions, and the feature flows are curves in two dimensions.
    The green cluster and red cluster contain input data points and the targets, respectively.}\label{fig: 01}
\end{figure}

 Consider the forward propagation of a  DNN with $L$ layers $\{x_l\}_{l=0,1,\dots,L}$:
\begin{equation}\label{eq: layer}
    x_{l+1} = h_l (x_l, w_l),
\end{equation}
where $x_{l+1}$ is the output feature of the $l$-th layer,
$h_l$ is the mapping in the $l$-th layer, and $w_l$ is the collection of trainable parameters.
Introducing a temporal partition: $\{t_l= l/L\}_{l=0}^L$ with time interval $\Delta t =1/L,$ and regarding $x_l$  as the value of a function $x(t)$ at time step $t_l$,
 without considering dimensional consistency, Eq.~(\ref{eq: layer}) can be rewritten as~\citep{resnet,Ewn,Dongbin,neural_ode}
\begin{equation}\label{eq: disc}
      x(t_{l+1}) = x(t_l) + \Delta t \cdot \hat{h}_l ( x(t_l), w_l),
\end{equation}
where $ \hat{h}_l = ( h_l - x_l)/ \Delta t $.
This can be interpreted as a discretization of evolution along a trajectory of the  network described by an ordinary differential equation~\citep{Ewn,Dongbin,neural_ode}
\begin{equation}\label{eq: ode}
    \frac{d x(t)}{dt} = \hat{h}(x(t), w(t), t).
\end{equation}
The feature flow is the  trajectory formed by connecting  features of  $\{x_l\}$, and is denoted by   $\Gamma{\{x_l\}}$. See Figure~\ref{fig: 01}a for an illustration of the feature flow.


We regard the trajectory $x(t)$ as well as the feature flow trajectory  $\Gamma{\{x_l\}}$ as a "curve".
Recall that for a curve $\gamma(t):\left[0,1 \right] \rightarrow R^D$  with arc length parameter $s$,
its length $ C(\gamma)$ and total absolute curvature $\mathcal{K}(\gamma)$ are
\begin{equation}
    C(\gamma):= \int_0^1 \|\gamma'(t) \| dt , \    \ \ \
    \mathcal{K}(\gamma):= \int_0^{C(\gamma)} |\kappa(s) | ds,
\end{equation}
where $\kappa(s)=\|\gamma''(s)\|$ is the curvature of the curve.

We introduce \textbf{feature flow regularization (FFR)} to improve the structured sparsity of DNN,
borrowing the definitions of length and total absolute curvature of a curve to the feature flow, i.e., the  trajectory formed by connecting  features of hidden layers.
For a feature flow associated with hidden layer features $\{x_l\}_{l=0,1,\dots,L}$,
the FFR is
\begin{equation}\label{eq: ffr_t}
    \mathcal{R}(x):= k_1 C(x) + k_2 \mathcal{K}(x),
\end{equation}
where
\begin{flalign} \label{eq: length}
    C(x) = \sum_{l=0}^{L-1} \| x_{l+1} - x_l\|, \ \ \ \
    \mathcal{K}(x)=\sum_{l=1}^{L-1} \| x_{l+1} -2 x_l + x_{l-1}\|, 
\end{flalign}
with $\| \cdot \| $ being the $L_1$ norm, and $k_1, k_2 >0$ the  hyperparameters.
Here up to some constant factors, $C(x) \sim
    \sum_{l=0}^{L-1} \| x'(t_l)\|$ is an approximation of the total length of the trajectory $x(t)$, and
$\mathcal{K}(x)\sim \sum_{l=1}^{L-1} | \kappa(t_l)|$ is an approximation of its total absolute curvature,
where  $\{x_l = x(t_l)\}$ is a discretization of trajectory $x(t)$ with time partition $\{t_l= l/L\}_{l=0}^L$.

FFR smooths the feature flow by controlling the length and curvature. Intuitively, the length term in Eq.~(\ref{eq: length}) makes the feature flow short,
and the curvature term in Eq.~(\ref{eq: length}) keeps the feature flow from bending too much.
As a result, DNN trained under FFR has a more sparse structure; See demonstration in 
Figure~\ref{fig: 01}b.  More quantitative analysis is given in Sec.~\ref{sec: Sparsity analysis}. We further give an illustration example in
Figure~\ref{fig: 01}c, which is a two-dimensional visualization showing the smooth effect of FFR on the feature flow of a ResNet. In this example, the input, features and output are all points in two dimensions, and the feature flows are actual curves in two dimensions.
We can see that the feature flow under FFR is shorter and straighter. More detail of this illustration example is given in Appendix Sec.~\ref{sec: example}.


\subsection{FFR applied to DNN and  pruning}
For a DNN and training dataset $\left\{ (x^{(j)},y^{(j)}\right\}_{j=1}^N$,
using Eq.~(\ref{eq: ffr_t}), the FFR is:
\begin{equation}\label{eq: ffr}
    \mathcal{R}(\displaystyle \sX ^{(j)} )
    = k_{1} \sum_{i=0}^{L-1} \|x_{l+1}^{(j)}- x_{l}^{(j)} \|
    +k_{2} \sum_{i=1}^{L-1}
    \|x_{l+1}^{(j)}-2 x_{l}^{(j)}+ x_{l-1}^{(j)} \|,
\end{equation}
where $\displaystyle \sX ^{(j)} =\left\{ x_l, l=0,\dots,L \right\}^{(j)}$ denotes the set of output features of hidden layers for the $j$-th input data.
The loss function with FFR in training is:
\begin{equation}\label{eq: loss}
       \frac{1}{N} \sum_{j=1}^{N} \left[ \mathcal{J} (x^{(j)},y^{(j)}, W)
       +\mathcal{R}(\displaystyle \sX ^{(j)} ) \right],
\end{equation}
where $\mathcal{J} (x,y, W)$ is the loss function before applying FFR.

Note that two hidden states $x_{l}, x_{l+1}$ may have different dimensions.
To fix the dimension mismatch problem,
we adopt the same strategy as \cite{resnet}, i.e, using a linear projection $\displaystyle \tP_{l }$ by the shortcut connections to match the dimensions.
We replace $x_{l}$ in Eq.~(\ref{eq: ffr}) by $\displaystyle \tP_{l} x_{l},$
where $\displaystyle \tP_{l} $ is the learnt projection matrix and will be treated as learnable parameters in training.
In implementation,
we first group the features according to the stage (features dimensions):
\begin{equation}\label{eq: group}
       \displaystyle \sX ^{(j)}
    =\bigcup_{g=1}^G \{ x_{g,1},x_{g,2},\dots,x_{g,l_g } \}^{(j)}  ,
    \    \
    L= \sum_{g=1}^G l_g.
\end{equation}
Here $G$ is the number of stages in $\displaystyle \sX ^{(j)}$ and $l_g$ is the number of hidden layers in stage $g$.
Secondly, we use the projection matrix to link different groups
since the dimensional mismatch only occurs at the first feature of each stage.
Using this method, the FFR of $\displaystyle \sX ^{(j)} $ becomes:
    \begin{equation} \label{eq: ffrt}
    \begin{aligned}
    & \mathcal{R}(\displaystyle \sX ^{(j)} )
    = k_{1} \sum_{g=1}^G \left[ \|x_{g,1}^{(j)}- \displaystyle \tP_{g} x_{g-1,l_{(g-1)}}^{(j)} \|  +  \sum_{i=1}^{l_g -1} \|x_{g,i+1 }^{(j)}-x_{g,i}^{(j)} \| \right]
    \\
    &+ k_{2} \sum_{g=1}^G \left[
    \|x_{g,2}^{(j)}-2 x_{g,1}^{(j)}+ \displaystyle \tP_{g} x_{g-1,l_{(g-1)}}^{(j)} \|
    +\sum_{i= 2}^{l_g-1 }\|x_{g,i+1}^{(j)}-2 x_{g,i}^{(j)}+ x_{g,i-1}^{(j)} \| \right].
    \end{aligned}
    \end{equation}
In the meantime,
the hyperparameters $k_1, k_2$ may vary with the feature dimensions so that FFR can uniformly control the features at different stages.
In our experiments, we adjust $k_1, k_2$ to be inversely proportional to the scale of feature maps. Moreover, the FFR process can be generalized to the case where we choose features every several layers.
We can denote the selected hidden layers as $\displaystyle \sL = \{ l_i, i=0,1,\dots, m ,  m=\# \displaystyle \sL\}.$
Then we apply FFR to the feature flow that connects the features in the set $\displaystyle \sX ^{(j)} =\{ x_l, l \in \sL \}$.

After training, we conduct one-shot filter pruning: removing filters with small magnitude and removing channels in the next layer that convolve with the feature maps generated by the pruned filters.
Finally, we fine tune the pruned network for a few epochs.

Our FFR training and pruning method is summarized in Algorithm \ref{alg: ffr}.
\begin{algorithm}[!hbtp]
\caption{FFR Training and One-shot Pruning}
\label{alg: ffr}
\begin{algorithmic}
\REQUIRE
training dataset $\left\{ (x^{(j)},y^{(j)}\right\}_{j=1}^N,$
a neural network and  hyperparameters $k_1, k_2$.
\STATE \textbf{Pre-step 1:} group the features in $\displaystyle \sX =\{ x_l, l \in \sL \}$ according to the stage as in Eq.~(\ref{eq: group}),
\STATE \textbf{Pre-step 2:} write down FFR $\mathcal{R}(\displaystyle \sX ^{(j)} ) $ in Eq.~(\ref{eq: ffrt})
for each paired data $(x^{(j)},y^{(j)})$.
\STATE \textbf{Training step:}
train the network under loss function with FFR given in Eq.~(\ref{eq: loss}).
\STATE \textbf{Pruning step:}  remove filters and the corresponding channels in the trained model,
\STATE fine tune the pruned model for a few epochs.
\RETURN a compact neural network.
\end{algorithmic}
\end{algorithm}

\FloatBarrier
\subsection{Feature flow of VGGNet and ResNet}
In this subsection, we demonstrate how to construct feature flows of VGGNet~\citep{vgg} and ResNet~\citep{resnet}, which are two commonly used network architectures.
Similar construction can apply to other neural networks.

\begin{figure}[!hbtp]
	\centering
	\begin{subfigure}[t]{0.26\textwidth}
		\centering
		\includegraphics[width=0.6\textwidth]{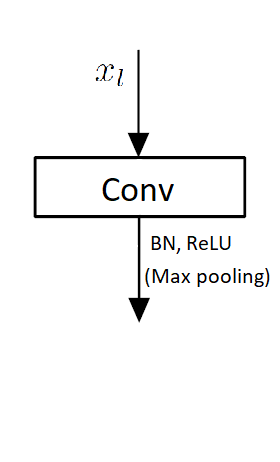}
		\caption{}
		\label{fig: vgg unit}
	\end{subfigure}
    \begin{subfigure}[t]{0.7\textwidth}
    	\centering
    	\includegraphics[width=0.6\textwidth]{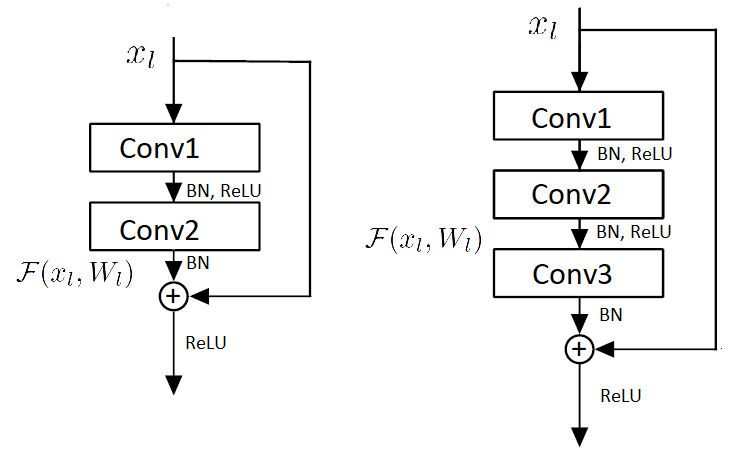}
    	\caption{}
    	\label{fig: residual unit}
    \end{subfigure}
    \caption{ (a) A convolutional block in VGGNet.
	(b) Residual blocks in ResNet.}
\end{figure}
\FloatBarrier
\paragraph{VGGNet.}
VGGNet is a convolutional neural network with a plain structure.
Each convolutional layer is followed by some activation functions: batch normalization, rectified linear activation function (ReLU), and sometimes a maxpooling layer.
We regard the convolutional layer with all its  activations as a block; see Figure \ref{fig: vgg unit}.
We collect the output of every convolutional block in VGGNet to form the feature flow $\Gamma{\{x_l\}}$.
Let $ x_{l}$ be the output feature of the $l$-th block,
then the feature of the $l+1$-th convolutional block is
\begin{equation}\label{eq: vggf}
    	x_{l+1} = (\text{Max pooling}) \cdot \sigma ( \mathcal{BN} \odot (W_l \otimes x_l)),
\end{equation}
where $W_l$ is the convolutional weights, $\mathcal{BN}$ is batch normalization, and $\sigma(\cdot)$ is ReLU.
Max pooling only appears once every several layers.
$ \otimes$ denotes convolution operation and $\odot $ denotes batch normalization operation.

\paragraph{ResNet.}
ResNet takes residual blocks as building blocks, each of which contains two or three convolutional layers; see Figure \ref{fig: residual unit}.
We collect the output of every residual block to form the feature flow $\Gamma{\{x_l\}}$.
The feature of $l+1$-th residual block is
\begin{equation}\label{eq: addition}
	x_{l+1} = \sigma( \mathcal{F}(x_{l}, W_{l}) + W_{s_l} x_{l} ),
\end{equation}
where $\mathcal{F}(x_{l}, W_{l})$ is the residual function,
$ W_{l} = \left\{ W_{l,k} | k=1,\dots,K \right\} $ is the set of convolutional weights, $K=2$ or $3 $ is the number of convolutional layers in each residual unit.
$ W_{s_l}$ is the identity matrix
if $x_{l}$ and $\mathcal{F}(x_{l}, W_{l})$ are in the same dimension,
and  the learnt projection matrix otherwise.

\FloatBarrier
\section{Sparsity analysis}
\label{sec: Sparsity analysis}
In this section, we demonstrate how  FFR improves structured sparsity.
As explained in the previous section, the idea of FFR is to shorten and straighten the trajectory of the input along the network. We will show that such effect
of FFR encourages feature sparsity in addition to penalizing the sparsity of the parameters.
Sparse features contribute to structured sparsity: for zero-value feature map, we can remove the filter which produces the feature map and the channel which convolves with the feature map.
Therefore, FFR improves the structured sparsity in the network.



\subsection{Feature and parameter sparsity}
\label{sec: ffr_analysis}
We analyze the effect of FFR when it is applied to the output features of the convolutional block and the residual block, which are commonly used in DNNs.
We focus on the length term in Eq.~(\ref{eq: length}) in FFR.

\begin{figure}[!htb]
\centering
\begin{tabular}{cc}
    \begin{minipage}[c]{0.56\textwidth}
    \centering
    \includegraphics[width=0.9\textwidth]{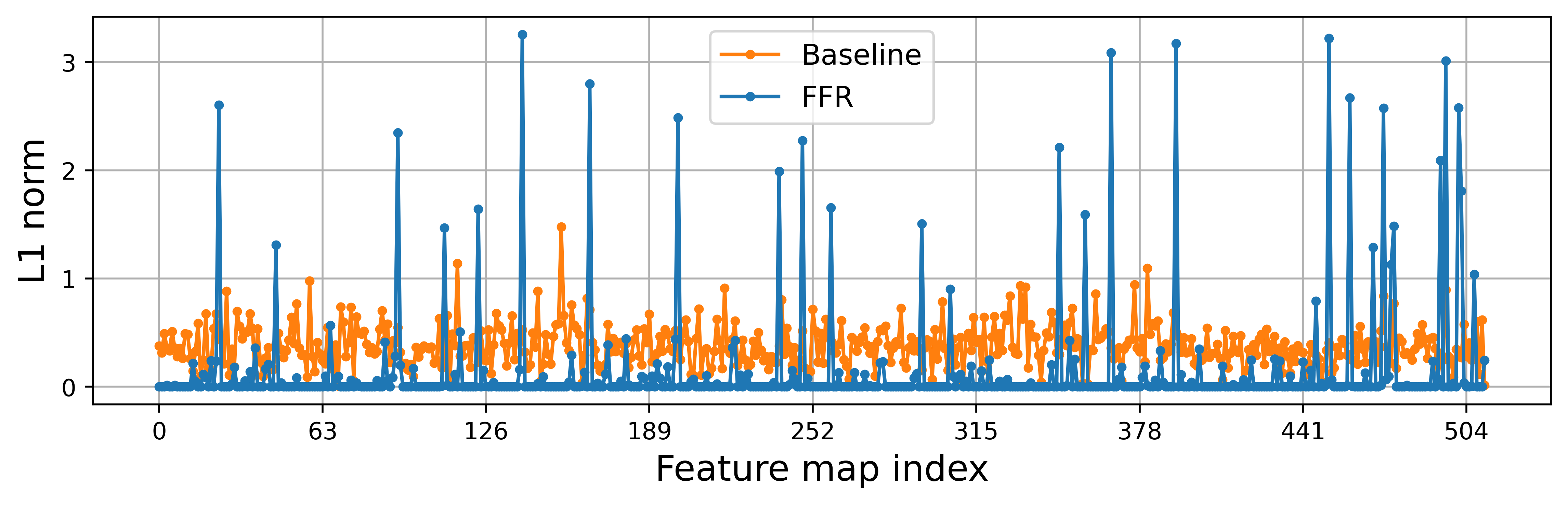}
    \subcaption{VGG16}
    \label{fig: fmaps}
    \includegraphics[width=0.9\textwidth]{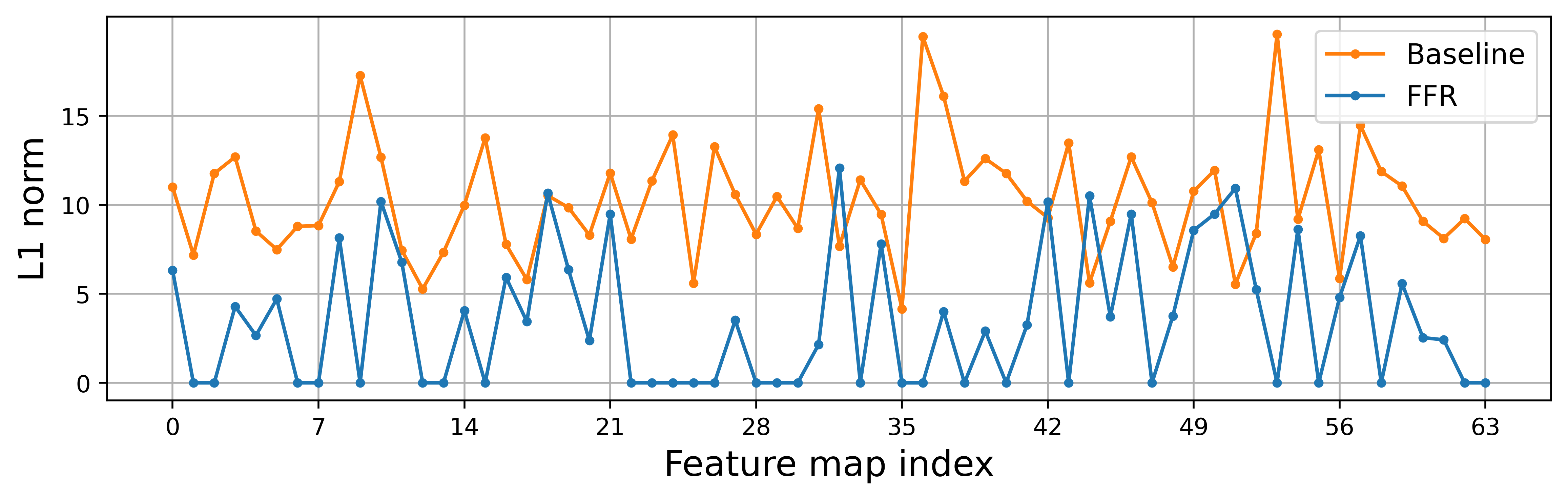}
    \subcaption{ResNet56}
    \label{fig: res_fmaps}
    \end{minipage}&
    \begin{minipage}[c]{0.42\textwidth}
    \centering
    \includegraphics[width= 1\textwidth]{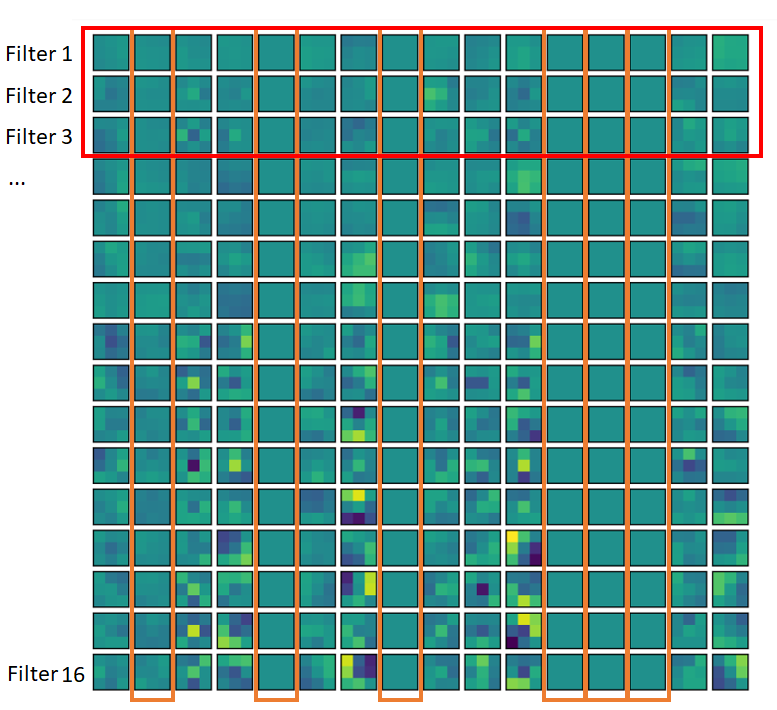}
    \subcaption{Filter visualization: $16\times16\times3\times3$.}
    \label{fig: filtersquare}
    \end{minipage}
\end{tabular}
\caption{
(a) VGG16 feature maps  trained with  and without FFR:  $L_1$ norm plot, with
512 feature maps in the feature of the last convolutional layer.
(b) ResNet56 feature maps trained with and without FFR:  $L_1$ norm  plot, with  64 feature maps in the feature of the nineteenth residual bock.
(c) Visualization of the filters in the first residual block in ResNet56  trained with FFR.
From top to bottom, the filters are displayed according to their norm from small to large. Square denotes the filters (red, horizontal) and channels (orange, vertical) that have small magnitude and can be removed.
Both VGG16 and ResNet56 are trained on CIFAR-10.}
\end{figure}

\paragraph{Convolutional block.}
Ignoring the activation functions in Eq.~(\ref{eq: vggf}), the length term in FFR is reduced to
\begin{equation}
    \| x_{l+1}-x_{l} \|
    = \| W_l \otimes x_l -x_l \|
    = \| (W_l-I) \otimes x_l \|,
\end{equation}
where $I$ denotes the 'identity' convolution kernels: for the $i$-th filter, only the $i$-th channel is a nonzero matrix and all other channels are zero-matrices.
This length term pushes the parameter $W_l$ to be close to the identity kernel $I,$ which is highly structured sparse.
In the meantime, the length term also pushes the feature $x_l$ to be sparse.
Figure \ref{fig: fmaps} shows the $L_1$ norm of 512 feature maps generated by the last convolutional layer in VGG16 trained with and without FFR on CIFAR-10. As shown in the figure, the baseline network (VGG16 trained without FFR) has  few zero-valued feature maps.
In contrast,
the network trained under FFR learns much more sparse features: most feature maps have zero-valued norm that can be removed. Such significant improvement of feature sparsity under FFR is also observed in the comparison for other layers of VGG16; see Appendix section~\ref{sec: norm_fm} for more plots.

\paragraph{Residual block.}
The length term in FFR enhances sparsity of the residual block by pushing the residual function $ \mathcal{F}(x_{l}, W_{l}) $
and features $x_{l} $ to zeros.
We ignore the projection matrix $W_{s} $ in the shortcut connection since it only appears when the stage changes.
From Eq.~(\ref{eq: addition}), the length term in FFR is reduced to
\begin{equation}
       \| x_{l+1}- x_{l} \|=
       \| \sigma( \mathcal{F}(x_{l}, W_{l}) + x_{l}) - x_{l} \|
       \stackrel{\text{element-wise}}{=}
       \begin{cases}
              \| \mathcal{F}(x_{l}, W_{l}) \| , & \mathcal{F}(x_{l}, W_{l}) + x_{l} \geq 0 ,\\
              \| x_{l} \| , &
              \mathcal{F}(x_{l}, W_{l}) + x_{l} < 0 .\\
       \end{cases}
\end{equation}
Under the hypothesis that the optimal function is closer to an identity
mapping than to a zero mapping \citep{resnet}, the first case
$\mathcal{F}(x_{l}, W_{l}) + x_{l} \geq 0$
holds most of the time, and in this case, FFR penalizes the norm of the residual function.
For a residual block with two or three convolutional layers, and ignoring the activations, the residual function can be reduced to
\begin{equation}
       \mathcal{F}(x_{l}, W_{l}) =    \sigma ( \mathcal{BN}_2 \odot W_{l,2} \otimes  \sigma (\mathcal{BN}_1 \odot W_{l,1} \otimes x_{l} ))
       \sim W_{l,2} \otimes W_{l,1} \otimes x_{l}, \ \text{or}
\end{equation}
\begin{equation}
       \mathcal{F}(x_{l}, W_{l}) =
       \sigma ( \mathcal{BN}_3 \odot W_{l,3} \otimes
       \sigma ( \mathcal{BN}_2 \odot W_{l,2} \otimes
       \sigma ( \mathcal{BN}_1 \odot W_{l,1} \otimes x_{l})))
       \sim
        W_{l,3} \otimes W_{l,2} \otimes W_{l,1} \otimes x_{l} .
\end{equation}
Hence imposing penalty on the residual function can make both parameters and features sparse.
In the second case, where $\mathcal{F}(x_{l}, W_{l}) <- x_{l} $, FFR encourages the feature $x_l$ to be sparse.

Figure \ref{fig: res_fmaps} shows the $L_1$ norm of 64 feature maps of the first residual block in ResNet56 trained with and without FFR on CIFAR-10. As shown in the figure,
the network trained under FFR outputs more zero value feature maps than
the baseline (trained without FFR).
In Figure \ref{fig: filtersquare}, we visualize the parameters of shape $16\times16\times3\times3$
from the first convolutional layer in the first residual block of ResNet56 trained with FFR on CIFAR-10.
We display the filter in row according to the magnitude: from top to bottom, the norm of the filter is increasing.
It can be seen from the figure that the parameters of the network trained with FFR are of high structured sparsity.
In the figure, small value filters are circled with a red horizontal square,
and  small value channels convolved with the same feature map are respectively circled with orange vertical squares.

\FloatBarrier

\subsection{Structured sparsity improvement}
\label{sec: Structured sparsity improvement}
\begin{figure}[!hbtp]
       \centering
       \begin{subfigure}[t]{0.45\textwidth}
              \centering
        \includegraphics[width=0.9\textwidth]{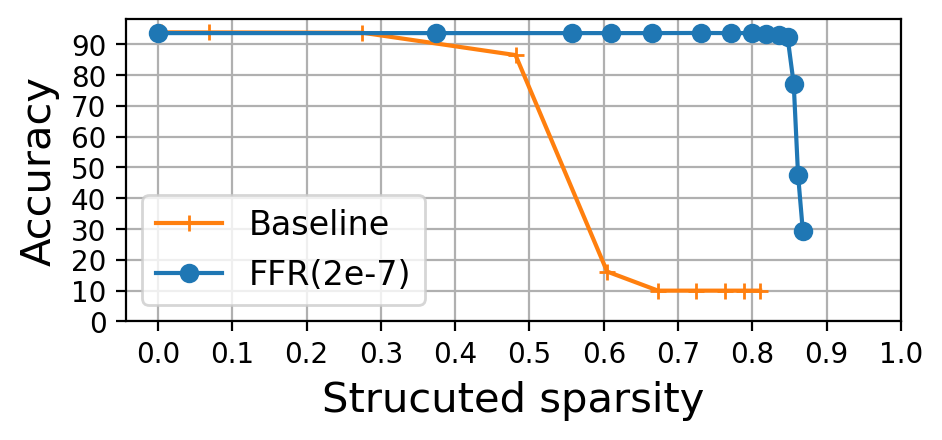}
           \caption{VGG16}
           \label{fig: vgg_sparsity}
       \end{subfigure}
    \begin{subfigure}[t]{0.45\textwidth}
       \centering
        \includegraphics[width=0.9\textwidth]{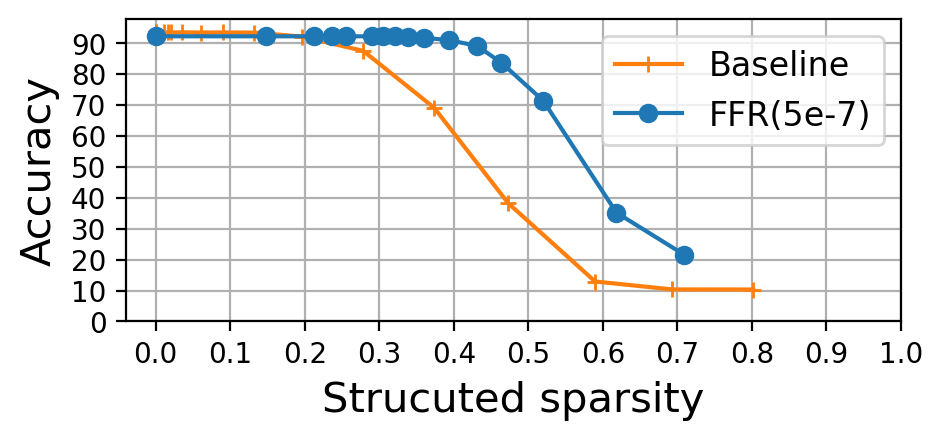}
        \caption{ResNet56}
       \label{fig: res_sparsity}
    \end{subfigure}
    \caption{ Accuracy-sparsity trade off curve of (a) VGG16 and
(b) ResNet56 trained with FFR (blue) and without FFR (orange, Baseline) on CIFAR-10.}
\end{figure}

To further show the ability of FFR in improving structured sparsity,
we examine the relation between accuracy and structured sparsity.
We compare VGG16 and ResNet56 with and without FFR trained on CIFAR-10.
After training, we use an increasing threshold to zero the filters whose magnitude is under the threshold and the corresponding channels. Then we calculate the accuracy and structured sparsity that is defined as the percent of the parameters zeroed.
Figure \ref{fig: vgg_sparsity} and \ref{fig: res_sparsity} show the accuracy-sparsity trade-off curves of VGG16 and ResNet56 on CIFAR-10, respectively.
 It can be seen that the network trained with FFR will not suffer accuracy degradation until it has a significantly large structured sparsity compared with the baseline case.

\section{Experiments}
\label{sec: experiment}

\subsection{Implementation}
\label{sec: Implementation}
\paragraph{Models and datasets.}
To demonstrate the effectiveness of FFR in pruning, we perform experiments with VGGNet and ResNet on the CIFAR-10 \citep{Krizhevsky09} and ImageNet \citep{imagenet} datasets.
The VGG16 is a modified VGGNet in \cite{vgg}: we remove two fully-connected layers and only use one fully-connected layer for classification.
We compare the network trained with FFR and without FFR as a baseline.
We adopt weights initialization as in \citet{weights}.

\paragraph{Training setting.}
FFR and baseline have the same training setting.
For CIFAR-10, the mini-batch size is 128.
We train VGG16 and ResNet56 200 epochs and set initial learning rate as 0.1 and divide it by 10 at the 80th, 120th, and 160th epoch.
For ImageNet, the mini-batch size is 256.
We train ResNet50 90 epochs and divide initial learning rate 0.1 by 10 at the 30th and 60th epoch.
We use SGD optimizer with weight decay $5e^{-4}$ for CIFAR-10, $1e^{-4}$ for ImageNet,  and momentum 0.9.
Note that $L_2$ regularization is implemented by applying weight decay in SGD.
We ran our experiments with pytorch.

\paragraph{Pruning setting.}
We perform one-shot structured pruning:
remove filters based on the magnitude and their corresponding channels in the next layer.
For all experiments on CIFAR-10 and ImageNet, we only fine tune the pruned model 30 epochs with learning rate $1e^{-4}$ on CIFAR-10 and $1e^{-3}$ on ImageNet.
For VGGNet, we prune the network directly, and
for ResNet, we use zeros padding in the pruned dimension.
The outputs of the shortcut and the last convolutional layer in the residual block in ResNet must have the same dimension because of the addition operation between them.
Using zero padding, we can remove the filters flexibly.
Moreover, if all filters of the first convolutional layer are pruned, then the residual block can be pruned. 

\subsection{Structured pruning}
\label{sec: Structured pruning}
\begin{figure}[!hbtp]
	\centering
    \includegraphics[width=0.9\textwidth]{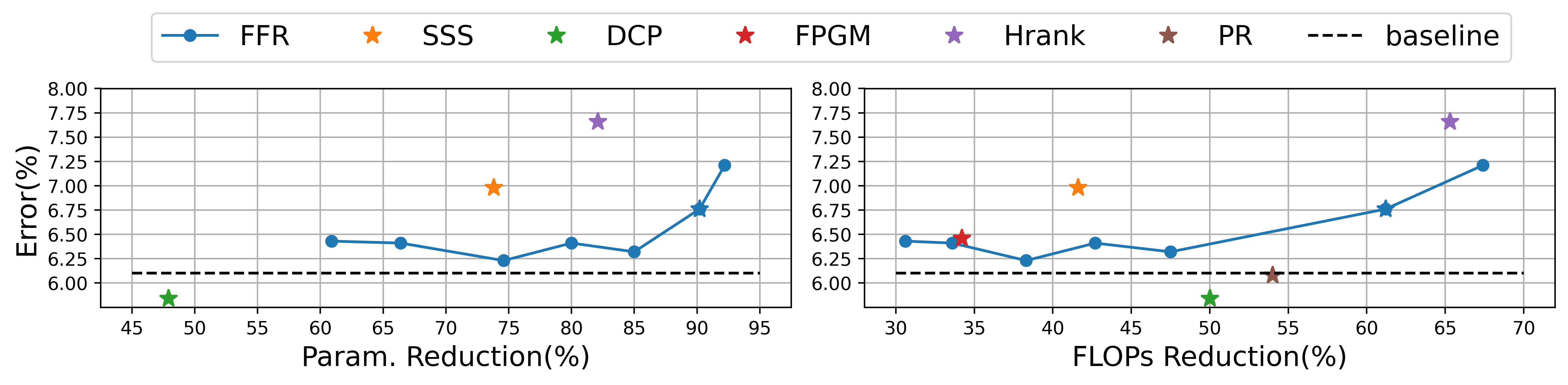}
    \caption{    Error-parameter reduction  and error-FLOPs reduction  trade-off curves of VGG16 trained with FFR on CIFAR-10, and comparison with the results of SSS~\citep{Huang}, DCP~\citep{DCP}, FPGM~\citep{FPGM}, Hrank~\citep{hrank}, PR~\citep{Polarization}.}
    \label{fig: pf_reduction}
\end{figure}

We compare pruning results of FFR with recent state-of-the-art pruning methods in
Figure~\ref{fig: pf_reduction} and Tables \ref{table: cifar10} and  \ref{table: imagenet}.
We present ablation study on the hypermeters $k_1, k_2$ in Appendix \ref{sec: ablation_study}.

\paragraph{Results on CIFAR-10.}
Table \ref{table: cifar10} shows the pruning results comparisons of VGG16 and ResNet56 on CIFAR-10.
For better comparison, in Figure \ref{fig: pf_reduction}, we plot error-parameter reduction
and error-FLOPs reduction trade-off curves of VGG16 trained with FFR together with the results of other methods shown in
Table \ref{table: cifar10}.
 We can see that FFR is able to achieve larger pruning ratio in terms of parameters and FLOPs. Specifically,
we prune VGG16 with 90.2$\%$ parameters reduction and 61.2$\%$ FLOPs reduction with a slight drop (0.66$\%$) in accuracy, and
  ResNet56 at a large FLOPs reduction 60.6$\%$ with accuracy drop 1.05$\%$.
From the comparisons of VGG16 results in Figure \ref{fig: pf_reduction}, the FFR method has lower error reductions in parameter and FLOPs than most of the available methods, and achieves larger pruning ratio.
Note that the DCP method~\citep{DCP} conducts channel pruning and fine tunes the network stage by stage to achieve lower error, whereas in FFR, we perform only one-shot structured pruning with a few epochs of fine tune.
%
We also visualize the feature maps of the reserved and pruned filters of the first convolution layer in VGG16 using FFR in Appendix \ref{sec: fm_v}, Figure \ref{fig: fm_visualization}.

\FloatBarrier
\paragraph{Results on ImageNet.}
Table \ref{table: imagenet} shows the pruning results comparisons of ResNet50 on ImageNet, which is a large-sacle dataset.
Although there is only one-shot pruning and a few epochs fine-tuning in FFR,
our FFR pruned ResNet50 achieves comparable parameters and FLOPs reductions with the recently proposed pruning methods.

\FloatBarrier
\paragraph{Methodology comparison.}
Methodologically, our FFR method manipulates the trajectory connecting features of hidden layers, which is different from pruning methods  based on feature maps information, e.g. \cite{Trim, FeatureS, hrank}
and is different from methods that impose regularization the parameters, e.g. \cite{Huang, Polarization}.
In terms of implementation simplicity, we use SGD during training and adopt a one-shot pruning strategy instead of additional optimization steps (e.g., \cite{FPGM}) or iterative pruning~\citep{DCP}.

\begin{table}[!hbtp]
	\caption{Pruning results on CIFAR-10. '-' represents result not reported. 'M' represents $1e^{6}$.}
	\label{table: cifar10}
	\centering
	\begin{tabular}{llcccc}
		\toprule
		Model & Method & Baseline & Pruned & Param. & FLOPs \\
		&    & Error ($\%$) &  Error ($\%$)  & Reduction & Reduction \\
		\midrule
		VGG16 & Baseline (Ours) & 6.10 & 6.10 & 0$\%$ (14.72M)  & 0$\%$ (313M) \\
		\cmidrule{2-6}
		& SSS~\citep{Huang} & 6.10 & 6.98 & 73.8$\%$  &   41.6$\%$  \\
		& DCP~\citep{DCP} &  6.01 & 5.84 &  47.9$\%$ &  50.0$\%$    \\
		 & FPGM~\citep{FPGM} & 6.42 & 6.46 & - & 34.2$\%$    \\
		& Hrank~\citep{hrank}  & 6.04 & 7.66 & 82.1$\%$  &  65.3$\%$     \\
		& PR~\citep{Polarization}  & 6.12  & 6.08 & -  &  54.0$\%$     \\
		& FFR $(k_1,k_2=2e^{-7})$ & 6.10 & 6.76 & 90.2$\%$ & 61.2$\%$   \\
	    \midrule
	    ResNet56 & Baseline (Ours) & 6.60 & 6.60 & 0$\%$ (0.86M) & 0$\%$ (126M) \\
		\cmidrule{2-6}
		& CP~\citep{SunJian}  & 7.20 & 8.20 &-  &   50.0$\%$   \\
	    & DCP~\citep{DCP} &  6.20 & 6.51 &  49.2$\%$ &  49.7$\%$    \\
	    & SFP~\citep{he2018soft}  & 6.41 & 6.65 &  -  &52.6$\%$     \\
	    & FPGM~\citep{FPGM} & 6.41 & 6.51 & - & 52.6$\%$   \\
		& Hrank~\citep{hrank}  & 6.74 & 6.83 & 42.4$\%$ & 50.0$\%$  \\
		& PR~\citep{Polarization}  & 6.20  & 6.17 & -  &  47.0$\%$    \\
		& FFR $(k_1,k_2=1e^{-7})$  & 6.60 & 7.65  &  $49.7\% $ &  $60.6\%$     \\
		\bottomrule
	\end{tabular}
\end{table}
\FloatBarrier

\begin{table}[!hbtp]
	\caption{Pruning results on ImageNet. '-'represents result not reported. 'M' represents $1e^{6}$.
	}
	\label{table: imagenet}
	\centering
	\begin{tabular}{lp{39mm}cccc}
		\toprule
		Model & Method   & Top-1 Acc.  & Top-5 Acc. & Param.  & FLOPs    \\
		 &    & Drop ($\%$) & Drop ($\%$) & Reduction & Reduction \\
		\midrule
	    ResNet50 & Baseline (Ours) & 0 (71.56) & 0 (90.28) & 0$\%$ (25.56M) & 0$\%$ (4089M) \\
		\cmidrule{2-6}
		& SSS~\citep{Huang}   & 4.30 & 2.07 & 38.8$\%$ &   43.0$\%$   \\
	    & DCP~\citep{DCP}   & 1.06 & 0.61  & 51.5$\%$  &  55.6$\%$   \\
	   & SFP~\citep{he2018soft}  & 14.01 & 8.27 &   -  &41.8$\%$  \\
	   & FPGM~\citep{FPGM}  &  1.32 & 0.55  &  - & 53.5$\%$    \\
		& Hrank~\citep{hrank}  & 4.17 & 1.86 &  62.1$\%$ & 46.0$\%$  \\
		& PR~\citep{Polarization}  & 0.52 &  -  &  -  &  54.0$\%$   \\
		& FFR $(k_1,k_2=5e^{-8})$    & 2.68  & 1.40  &  52.3$\% $ &  47.2$\%$ \\
		\bottomrule
	\end{tabular}
\end{table}

%

\FloatBarrier
\section{Conclusions and discussion}
In this paper, we propose a simple and effective regularization method (FFR)
from a new perspective of the data trajectory along the network.
FFR smoothes the trajectory by imposing controls on the length and total absolute curvature of the feature flow,
leading to significant increase of structured sparsity in DNNs.
We perform a sparsity analysis of FFR for VGGNet and ResNet to validate  this method.
Experimental results show that FFR can significantly enhance structured sparsity, which enables us to prune filters efficiently in one pass.
Future work may include:
(1) perform rigorous sparsity analysis for FFR, and
(2) integrate FFR with other pruning methods to further improve pruning results.
Since in FFR, the regularization is imposed on features,
  it is not easy to predetermine the sparsity, although experimental results using FFR have shown significant sparsity improvement.
In order to balance  sparsity and accuracy, we need to tune the hyperparameters $k_1$, $k_2$, which vary with  networks and datasets. Further improvements in these aspects will also be explored in the future work.



\FloatBarrier

\newpage



\bibliography{iclr2022_conference}
\bibliographystyle{iclr2022_conference}

\newpage
\appendix
\section{Appendix}
\subsection{Setting of the illustration example in section \ref{sec: ffr}}
\label{sec: example}

To visualize the smoothing effect of FFR, we use a two-dimensional example.
In Figure \ref{fig: 01}c,
the green and red clusters contain 50 paired data points in two dimensional space.
The green cluster is evenly distributed in a circle with center $(2,6)$ and radius $0.5$.
The red cluster is obtained by exactly shifting the green cluster four units to the right and four units down, that is in a circle with center $(6,2)$.
we use a ResNet to learn a translation mapping from the green cluster to the red cluster.
The ResNet has five fully-connected residual blocks.
Here each residual block contains two linear layers and outputs a two-dimensional feature.
We train ResNet without and with FFR to learn the mapping under the same setting for comparison.
A trajectory is obtained by starting from an input data point,  connecting five output features and ending with the output.
From Figure \ref{fig: 01}c, the trajectories of the network trained with FFR, represented by the orange curves,
are shorter and more straight than those of the network trained without FFR, represented by the blue curves.
We plot the feature flows of three test data for comparison.
It shows that FFR indeed effectively smooths the feature flow.

\subsection{Feature maps norm plot in section \ref{sec: ffr_analysis}}
\label{sec: norm_fm}
\FloatBarrier

\begin{figure}[!hbtp]
       \centering
       \begin{subfigure}[t]{0.9\textwidth}
              \centering
              \includegraphics[width=1\textwidth]{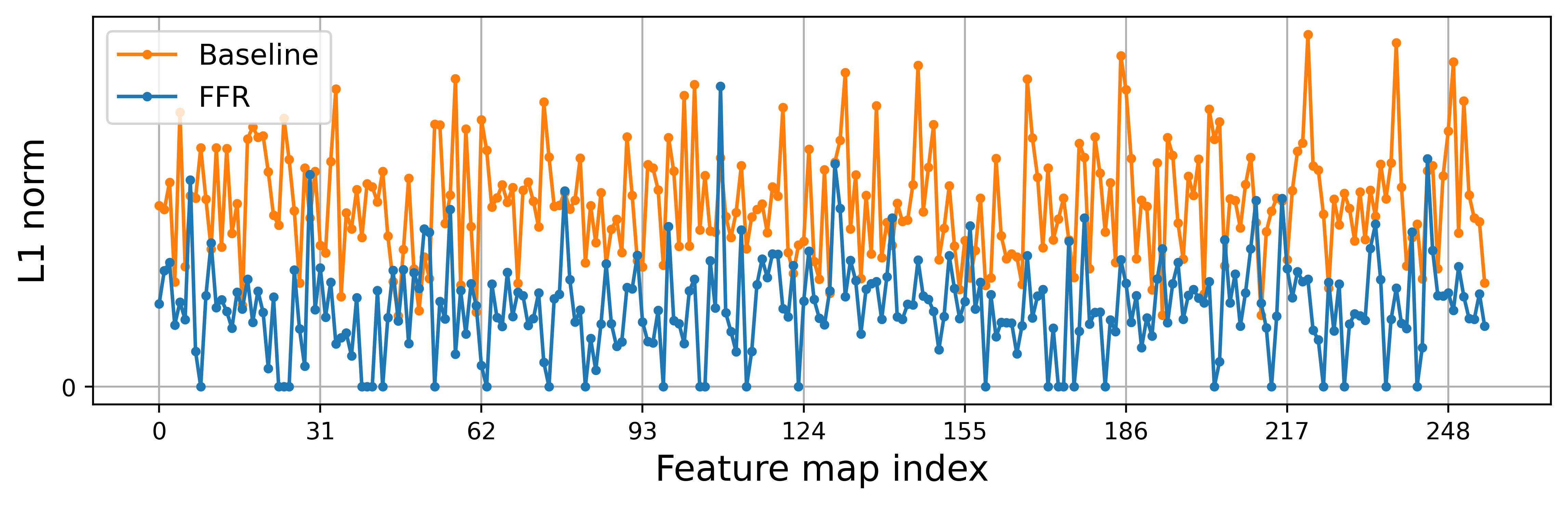}
              \caption{256 feature maps of the fifth convolutional layer.}
              \label{fig: fmaps_layer5}
       \end{subfigure}

       \begin{subfigure}[t]{0.9\textwidth}
              \centering
              \includegraphics[width=1\textwidth]{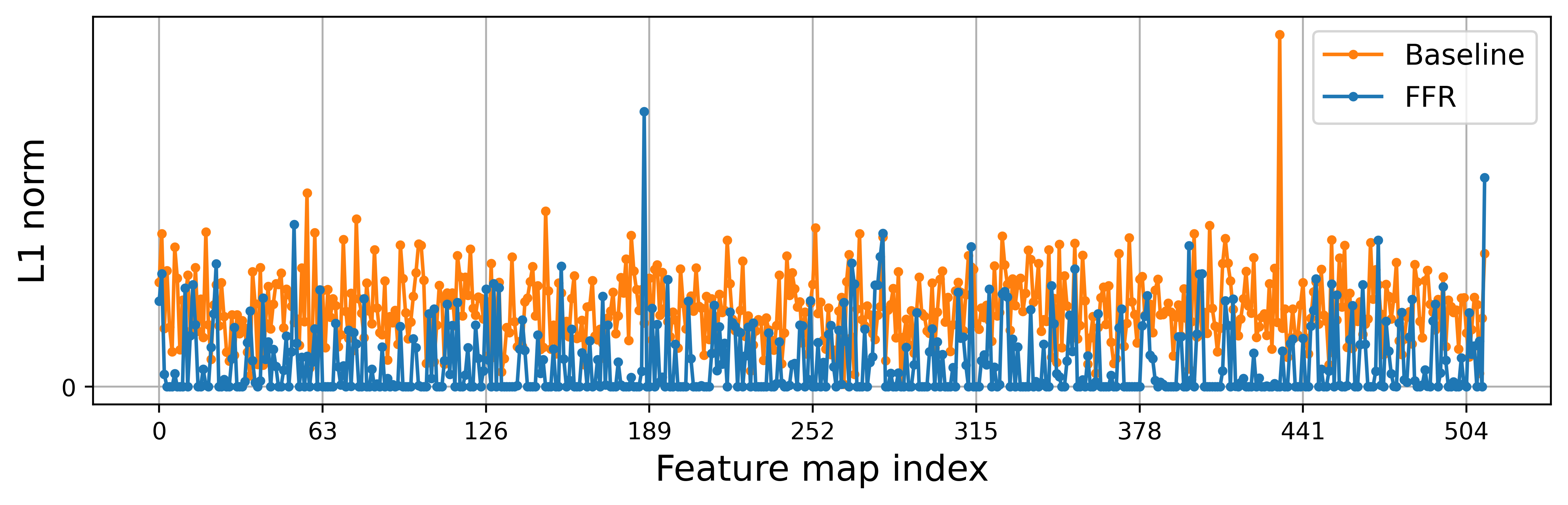}
              \caption{512 feature maps of the eighth convolutional layer.}
              \label{fig: fmaps_layer8}
       \end{subfigure}
    \caption{ VGG16 feature maps trained with and without FFR on CIFAR-10: $L_1$ norm plot.}
\end{figure}

We show the $L_1$ norm plot of 256 feature maps generated by the fifth convolutional layer and
512 feature maps generated
by the eighth convolutional layer in VGG16 trained with and without FFR on CIFAR-10 in Figure \ref{fig: fmaps_layer5} and \ref{fig: fmaps_layer8} respectively.
Consistent with the claim in the main text,
the network trained under FFR learns much more sparse features with
zero-valued norm than
the baseline network (VGG16 trained without FFR).
FFR improves the feature sparsity which contributes to the structures sparsity in DNN.

\newpage
\FloatBarrier
\subsection{Feature maps visualization}
\label{sec: fm_v}
\begin{figure}[!hbtp]
	\centering
	\begin{subfigure}[t]{0.17\textwidth}
		\centering
		\includegraphics[width=0.9\textwidth]{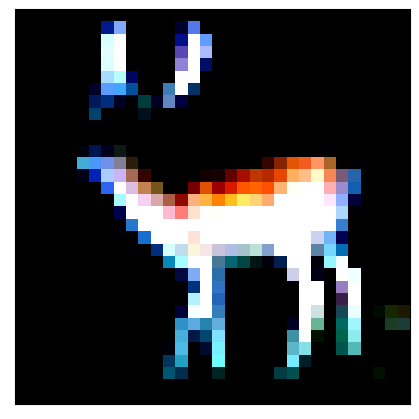}
		\caption{Input image}
		\label{fig: input}
	\end{subfigure}
    \begin{subfigure}[t]{0.4\textwidth}
    	\centering
    	\includegraphics[width=0.9\textwidth]{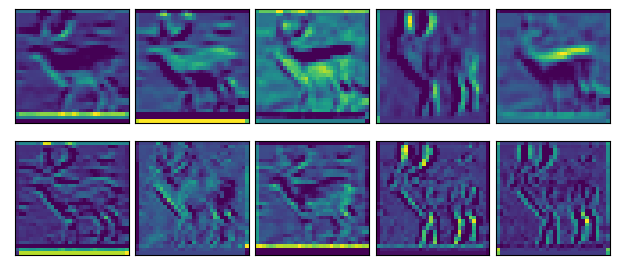}
    	\caption{Feature maps of preserved filters}
    	\label{fig: fm_remian}
    \end{subfigure}
    \begin{subfigure}[t]{0.4\textwidth}
    	\centering
    	\includegraphics[width=0.9\textwidth]{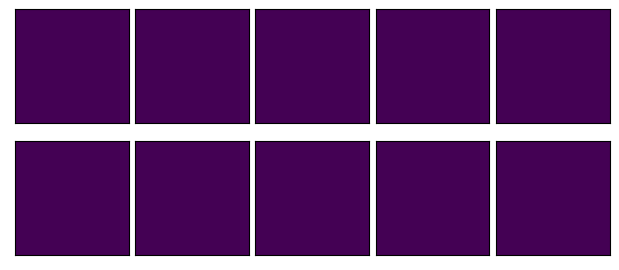}
    	\caption{Feature maps of pruned filters}
    	\label{fig: fm_pruned}
    \end{subfigure}
    \caption{ Feature maps visualization of the first convolutional layers in VGG16 using FFR.}
    \label{fig: fm_visualization}
\end{figure}

\FloatBarrier
\subsection{Ablation study}
\label{sec: ablation_study}
We tune the hyperparameters $k_1, k_2$ with experiments.
The choice of hyperparameters depends on the architecture of the DNN and the scale of the input: for the DNN with more layers and higher resolution input.
we use smaller $k_1, k_2$.
We train VGG16 under FFR with different hyperparameters on CIFAR-10 and conduct pruning experiments.

We list the pruning results of VGG16 on CIFAR-10.
The parameter reduction, FLOPs reduction and accuracy data are obtained by using the  same threshold to prune VGG16 trained under FFR with different hyperparameters $k_1, k_2$ and fine tuning the pruned model 30 epochs.

As shown in Table \ref{table: ablation}, the baseline (VGG16 trained without FFR) cannot preserve the accuracy under the same pruning and fine-tuning setting.
Moreover, network trained under FFR with appropriate $k_1, k_2$ achieves large pruning ratio while maintaining the acurracy.

\begin{table}[!hbtp]
       \caption{Alation study on $k_1, k_2$.
       Pruning results of VGG16 on CIFAR10. 'M' represents $1e^{6}$.
       }
       \label{table: ablation}
       \centering
       \begin{tabular}{lccc}
              \toprule
              Hyperparamerer   &  Error($\%$)   & Param. Reduction & FLOPs Reduction\\
              \midrule
           Baseline & 6.10 & 0$\%$(14.72M)  & 0$\%$(313M)  \\
           $k_1, k_2=0$  & 34.57 & 74.1$\%$  & 35.0$\%$ \\
           $k_1, k_2=5e^{-8}$ & 6.36  & 76.5$\%$  &   36.4$\%$  \\
           $k_1, k_2=1e^{-7}$ & 6.45  & 77.5$\%$  &   38.6$\%$  \\
           $k_1, k_2=2e^{-7}$ & 6.32  & 85.0$\%$  &   47.5$\%$  \\
              \bottomrule
       \end{tabular}
\end{table}

\end{document}